%% file: main.tex
\documentclass[a4paper,10pt]{article}

\usepackage[pdftex]{graphicx}
\DeclareGraphicsExtensions{.pdf,.png,.jpg}

\usepackage{amsmath}
\usepackage{amsthm}
\usepackage{array}
\usepackage{multirow}
\usepackage{enumerate}
\usepackage{rotating}
\usepackage{amssymb}
\usepackage{setspace}
\usepackage{epsf}
\usepackage{epsfig}
\usepackage{lscape}
\usepackage{harvard}
\pdfoutput=1

\title{Acceleration of a procedure to generate fractal curves of a given dimension through the probabilistic analysis of execution time}
\author{Manuel Cebri\'an, Manuel Alfonseca and Alfonso Ortega\thanks{The
authors are with the Departamento de Ingenier\'ia Inform\'atica, Escuela
Polit\'ecnica Superior, Universidad Aut\'onoma de Madrid, 28049 Madrid,
Spain, fax number: (+34) 914 972 235, e-mail: \{manuel.cebrian,
manuel.alfonseca, alfonso.ortega\}@uam.es.}}

\date{}

\theoremstyle{definition}

\begin{document}

\maketitle

\input{abstract}


\input{algorith}
\input{cap3}

\input{cap4-2}
\input{cap5}
\input{cap4-4-7}

\section*{Acknowledgements}
This work has been sponsored by the 
Spanish Ministry of Education and Science (MEC), project number 
TSI2005-08225-C07-06.

\bibliographystyle{harvard}
\bibliography{/home/mcebrian/investigacion/main.bib}

\end{document}

%% file: abstract.tex
\begin{abstract}
In a previous work, the authors  proposed a Grammatical Evolution algorithm to
automatically generate Lindenmayer Systems which
represent fractal curves with a pre-determined fractal dimension. This paper
gives strong statistical evidence that the probability distributions of the execution time of that 
algorithm exhibits a heavy tail with an hyperbolic probability decay for long executions, 
which explains the erratic performance of different executions of the algorithm. 
Three different restart strategies have been incorporated in the algorithm to
mitigate the problems associated to heavy tail distributions: the first
assumes full knowledge of the execution time probability distribution, the
second and third assume no knowledge. These strategies exploit the fact that 
the probability of finding a solution in short executions is non-negligible and yield a severe
reduction, both in the expected execution time (up to one order of magnitude) and in its
variance, which is reduced from an infinite to a finite value.
\end{abstract}

%% file: algorith.tex
\section{Introduction}

In the last decades, genetic algorithms, which emulate biological evolution in computer software,
have been applied to ever wider fields of research and development and have given rise to a few
astounding successes, together with a certain mount of disappointment, frequently related to
the apparently inherent slowness of the procedure. This is not a surprise, as biological evolution,
which serves as the source for most of the ideas used by the research in genetic algorithms,
makes a extremely slow and difficult to experiment field, where actual processes require
millions of years in many cases. This slowness is in part a consequence of the fact that
randomness is a basic underlying of the search performed by genetic algorithms.
For this reason, the discovery and proposal of procedures to accelerate their execution time
is one of the most interesting open questions in this field.

The procedure we propose in this paper has made it possible to increase by an order of
magnitude the performance of at least one application of genetic algorithms: the use of grammatical
evolution to generate fractal curves of a given dimension. It is probable that the application
of the same procedure may be useful to accelerate many other applications of similar techniques,
although there are cases where it cannot provide any improvement. The paper offers ways
to predict the situations where this procedure may be useful, and recognize those
where it will not provide any improvement, by analyzing the statistical distributions of the
execution time of the algorithms. In fact, the family of heavy-tail distributions embodies those
applications where the best improvement can be attained by the application of re-start
techniques, while another family (leptokurtic distributions) also offer a
significant
acceleration.

The remainder of this introduction contains a simple introduction to the three main
fields affecting the experiment we have used as the template for the experimentation
of the acceleration techniques: the family of genetic algorithms we are testing (grammatical evolution);
fractal curves and their dimension; and L systems, which provide an easy way to represent
the former and making their computation straightforward.

Section 2 summarizes an algorithm we have developed and described in a previous
publication, which makes it possible to compute the dimension of a fractal curve
from its equivalent L system. Section 3 describes the concrete case we have used as
the benchmark for our acceleration techniques: a genetic algorithm which generates a
fractal curve with a given dimension. This algorithm has also been previously published
in the scientific literature.

Section 4 describes the families of heavy tail and leptokurtic distributions,
where the acceleration techniques proposed in this paper are most useful. Section 5
proves that the experiment described in section 3 gives rise to execution time distributions
belonging to those families. Section 6 describes the restart strategy whose use
significantly
accelerates the execution time of our algorithm and all others with a distribution in the same
families. Finally, section 7 offers the conclusions of the paper and proposes several lines
of future work.

\subsection{Grammatical evolution}

Evolutionary Automatic Programming (EAP) refers to those systems that use
evolutionary computation to
automatically generate computer programs.
EAP techniques can be classified according to the
way the programs are represented: tree-based systems, which
work with the derivation trees of the programs, or string-based
systems, which represent them as strings of symbols. 
The best known tree-based system
is genetic programming (GP), proposed by \citeasnoun{Koza1992}, which
automatically generates LISP programs to
solve given tasks.

Tree-based systems do not make an explicit distinction between genotype and phenotype.
String-based systems may do it. Grammatical evolution 
\citeaffixed*{O'Neill2003a}{GE, } is the latest, most promising string-based
approach.
GE is an EAP algorithm based on strings, independent
of the language used. Genotypes are represented by strings of
integers (each of which is called a \emph{codon}) and the context-free
grammar of the target programming language is used to deterministically
map each genotype into a syntactically correct phenotype
(a program). In this way, GE avoids one of the main difficulties in EAP,
as the results of applying genetic operators to the individuals in a population are guaranteed
to be syntactically correct.
The following scheme shows the way in which GE combines
traditional genetic algorithms with genotype-to-phenotype
mapping.

\begin{enumerate}
\item A random initial population of genotypes is generated.
\item Each member of the population is translated into its corresponding phenotype.
\item The genotype population is sorted by their fitness (computed from the phenotypes).
\item If the best individual is a solution, the process ends.
\item The next generation is created: the mating-pool is chosen with
a fitness-proportional parent selection strategy;
the genetically modified offspring is generated, and the
worst individuals in the population are replaced by them.
\item Go to step 2.
\end{enumerate}

\begin{figure}[h]
\centering \includegraphics[width=300pt]{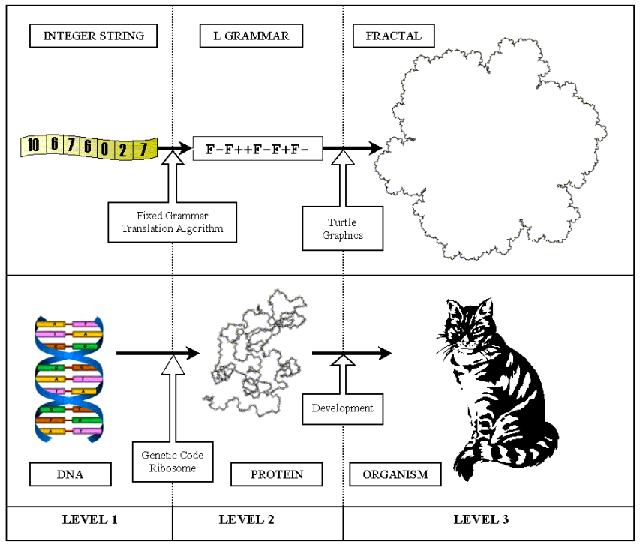}
\caption{Graphical scheme of a GE process}
\label{biological_scheme}
\end{figure}

This procedure is similar in many respects to biological evolution. There are
three different levels. Figure \ref{biological_scheme} shows a graphical scheme
of  the process in the particular case studied in this paper: the automatic 
generation of fractal curves with a given dimension.

\begin{itemize}
\item The \emph{genotype} (nucleic acids), is represented in GE by vectors of integers.
\item The \emph{intermediate} level (proteins), is represented in GE by
words in a given alphabet, which in our case describe an L system (see below).
The translation from the genotype to the
intermediate level is performed by means of a fixed grammar (the equivalent of the
fixed genetic code).
\item The \emph{phenotypic} (organisms), in our case represented by
the fractal curves obtained from the intermediate-level words by means of a graphical interpretation.
\end{itemize}

\subsection{Fractals and fractal dimension}

The concept of dimension is very old and seems easy and
evident:  sometimes it can be clearly and elegantly defined as the
number of directions in which movement is allowed: with this interpretation,
dimensions are consecutive integers: 0 (a point), 1 (a line), 2 (a surface),
3 (a volume), with no doubtful cases. This is called a \emph{topological dimension}.
However, as \citeasnoun {Mandelbrot83} describe in his seminal article, some
doubtful cases 
exist: depending on the size of the observer, a ball of thread can be considered
as a point (dimension 0, for a large observer), a sphere (dimension 3, for an observer comparable to the ball),
a twisted line (dimension 1, for a smaller observer), a twisted cylinder (dimension 2, for an even smaller observer),
and so forth.

There is a class of apparently one-dimensional curves for which the concept of dimension is tricky:
in 1890, Giuseppe Peano defined a curve which goes through every point in a
square, and therefore can be considered as two-dimensional. In 1904, Helge von Koch devised another, whose
shape reminds a snowflake and whose longitude is infinite, although the surface it encloses is limited.
Von Koch's snowflake does not fill a surface, therefore its dimension should be greater than 1 but less than 
2. In 1919, Hausdorff proposed a new definition of dimension, applicable to such doubtful cases:
curves such as those just described may have a fractional dimension, between 1 and 2. Peano's
curve has a Hausdorff dimension of 2; Von Koch's snowflake has a Hausdorff dimension of 1.2618595071429...
Other alternative definitions of dimension were proposed during the twentieth century,
such as the Hausdorff-Besicovitch dimension, the Minkowsky dimension, or the boxcounting
dimension \citeaffixed{falconer1990fgm,yamaguti1997mf}{see}. They differ
only in details and are known as \emph{fractal dimensions}.

The name \emph{fractal} was introduced in 1975 by Mandelbrot
and applies to objects with some special properties,
such as a fractal dimension different from their integer topological dimension, self-similarity (containing 
copies of themselves), and/or  non-differentiability at every point.

Fractal curves have been generated or represented by
different means, such as fractional Brownian movements,
recursive mathematical families of equations (such as
those that generate the Mandelbrot set), and recursive
transformations (generators) applied to an initial shape
(the initiator). They have found applications in antenna design,
the generation of natural-looking landscapes for artistic purposes,
and many other fields. The generation of fractals with a given 
dimension can be useful for some of these applications.

This paper discusses only the initiator-generator family of fractals.

\subsection{L systems}

L systems, devised by \citeasnoun{lindenmayer1968mmc},
also called parallel-derivation grammars, differ from
Chomsky grammars because derivation is not performed sequentially
(a single rule is applied at every step) but in parallel (every symbol is
replaced by a string at every step).
L systems are appropriate to represent fractal curves
obtained by means of recursive transformations \cite{culik1993sam}.
The initiator maps to the axiom of the L system;
the generator becomes the set of production rules;
recursive applications of the generator to the initiator
correspond to subsequent derivations of the axiom. The
fractal curve is the limit of the word derived from
the axiom when the number of derivations tends to
infinity.

Something else is needed: a graphic
interpretation which makes it possible to convert the
words generated by the L system into visible
graphic objects. Two different families of graphic interpretations of
L systems have been used: turtle graphics and vector
graphics. In a previous paper,
we have proved an equivalence
theorem between two families of L systems,
one associated with a turtle graphics interpretation,
the other with vector graphics \cite{alfonseca1997srf}. Our
theorem makes it possible to focus only on turtle graphics without a
significant loss of generality.

The turtle graphics interpretation was first
proposed by \citeasnoun{papert1980mcc} as the trail left by an invisible
\emph{turtle}, whose state at every instant is defined by its position and the
direction in which it is looking. The state of the turtle changes as it moves a
step forward or as it rotates by a given angle in the same position.
Turtle graphics interpretations may exhibit different
levels of complexity. We use here the following version:

\begin{itemize}
\item The angle step of the turtle is $\alpha=(2k\pi/n)$, where
$k$ and $n$ are two integers.
\item The alphabet of the L system is expressed as
the union of the four disjoint subsets: $N$ (non-graphic symbols), $D$ (visible graphic symbols,
which move the turtle one step forward, in the direction of its current angle, leaving a visible trail),
$M$ (invisible graphic symbols, which move the turtle one step forward, in the direction of its
current angle, leaving no visible trail) and extra symbols such as $\{+, -\}$, which increase/decrease
the turtle angle by $\alpha$, or a parenthesis pair, which are used in conjunction with a stack to add
branches to the images. These symbols usually are associated with L system trivial rules such as
$+ ::= +$. In the following, the trivial rules will be omitted but assumed to be present.
\end{itemize}

A string is said to be \emph{angle-invariant} with a turtle graphics interpretation
if the directions of the turtle at the beginning and the end of the string are the same.
In this paper we only consider \emph{angle-invariant D0L systems} (where D0L describes a
deterministic context-free L system), i.e. the set of
D0L systems such that the right-hand side of all of their rules is an
angle-invariant string.

Summarizing: a fractal curve can be represented by means
of two components: an L system and a turtle graphics
interpretation, with a given angle step. The length of the moving
step (the scale) is reduced at every derivation in the
appropriate way, so that the curve always occupies the
same space.

\section{An algorithm to determine the dimension of a fractal curve from its equivalent L system}

Several classic techniques make it possible to estimate the dimension of a fractal curve. All
attempt to measure the ratio between how much the curve grows in length, and how
much it advances. The ruler dimension estimation computes the dimension of a fractal curve as a
function of two measurements taken while \emph{walking} the curve in a number of discrete steps.
The first measurement is the \emph{pitch length} ($p_l$), the length of the step
used, which is
constant during the walk. The second is the number of steps needed to reach the end of the walk
by walking around the fractal curve, $N\left(p_l\right)$.
The fractal dimension, $D_{p_l}$, is defined as

\begin{equation}
D_{p_l} = \lim_{p_l \rightarrow
0^+} \frac{-\log N(p_l)}{\log p_l}  
\end{equation}

In a previous work \cite{Alfonseca01} we presented an
algorithm that reaches the same result
by computing directly from the L system that represents the fractal curve,
without performing any graphical representation.
The L system is assumed to be an angle-invariant D0L system with a single draw symbol.
The production set consists of a single rule,
apart from trivial rules for symbols $+$, $-$, $($, and $)$.
Informally, the algorithm takes advantage of the fact
that the right side of the only applicable rule provides a
symbolic description of the fractal generator, which can
be completely described by a single string. The
algorithm computes two numbers: the length
$N$ of the visible walk followed by the fractal generator
(equal in principle to the number of draw symbols in the
generator string), and the
distance $d$ in a straight line from the start to the endpoint
of the walk, measured in turtle step units (this number can
also be deduced from the string). The fractal dimension is:

\begin{equation}
D = \frac{\log N}{\log D}
\label{eq4.jpg}
\end{equation}

The scale is reduced at every derivation in such a way that
the distance between the origin and the end of
the graphical representation of the strings is always the
same. For instance, the D0L scheme associated with the rule

\begin{figure}
\centering 
\includegraphics[width=200pt]{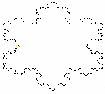}
\caption{Von Koch snowflake curve.}
\label{snowflake}
\end{figure}

$F ::= F+F--F+F$

with axiom $F--F--F$ and a turtle graphic
interpretation, where ${F}$ is a visible graphic symbol and the step
angle is 60, represents the fractal whose fifth derivation
appears in figure \ref{snowflake}.

The string $F+F--F+F$ describes the fractal generator. The number of
steps along the walk ($N$) is the number of draw symbols
in the string, 4 in this case. The distance $d$ between the
extreme points of the generator, computable from the
string by applying the turtle interpretation, is 3.
Therefore, the dimension is

\begin{equation}
 D = \frac{\log 4}{\log 3} = 1.2618595074129\ldots
\label{eq5.jpg}
\end{equation}

This is the same dimension obtained with other methods, as
specified in  \cite[p.~42]{Mandelbrot83}.

This algorithm can be easily extended to fractals whose L systems contain
more than one draw symbol and more than one rule, if all the rules
preserve the ratio between $N$ and $d$ in the previous expression.
Most of the initiator-iterator fractals found in the literature 
can be represented by angle-invariant D0L systems whose draw 
symbols-contribute to the dimension in this way.
The algorithm was also refined to successfully include fractal curves
which overlap, either in the generator itself, or after subsequent derivations.
In those cases, the definition of the fractal dimension is replaced by

% FIXME: el nivel de detalle aqui no es relevante para este articulo, haciendo
% que se parezca demasiado al articulo original (vuestro).

%\begin{itemize}
%\item The length $N$ of the visible walk may not be equal to
%the number of draw symbols in the generator string.
%This may happen for two reasons:
%\begin{itemize}
%\item The turtle graphic associated with the string may overlap, i.e. pass
%more than once along a set of points with a nonzero measure, as in the PD0L 
%scheme:

%$F::= F+FF+++F++F-FF+++F++F--F$

%with a step angle of 45 and axiom F++F++F++F.

%\begin{figure}
%\centering \includegraphics[width=\textwidth]{ldimen2f}
%\caption{An overlapping fractal curve}
%\label{LDIMEN2F.TIFF} 
%\end{figure}

%Figure \ref{LDIMEN2F.TIFF} shows the third derivation of this fractal curve. 
%Our algorithm has been refined to take this case into account in such
%a way that the appropriate value of $N$ is computed,
%where such sets of points are counted only once. This
%means that the value of $N$ may be noninteger, (in the example,
%its value is 8 plus the square root of 2).

%\item The turtle graphic associated with a derivation of the
%string may overlap, as in the PD0L scheme:

%$F ::= F+FF-F-FF+F$

%with a step angle of 90 and axiom F+F+F+F. 

%\end{itemize}
%\end{itemize}

%\begin{figure}
%\centering \includegraphics[width=\textwidth]{ldimen3f}
%\caption{Another overlapping fractal curve}
%\label{LDIMEN3F.TIFF} 
%\end{figure}

%Figure \ref{LDIMEN3F.TIFF}
%represents the generator of the corresponding fractal
%curve and its fourth derivation. In this case,
%the definition of fractal dimension is replaced by

\begin{equation}
D = \lim \frac{\log N}{\log d}
\label{eq6.jpg} 
\end{equation}
where the limit is taken when the number of derivations goes to infinity.
Our algorithm computes this case by
computing the dimension of a certain number of derivations until the
quotient converges. %The resulting dimension of the example is

\section{Grammatical evolution to design fractal curves with a given dimension}

Designing fractal curves with a given dimension is
relatively easy for certain values of the desired dimension (for instance,
1.261858... or $\log4/\log3$), but very difficult for others (the reader can try to
hand design a fractal curve with a dimension of 1.255). To do it, one has to
find two integer numbers, $a$ and $b$, such that 1.255 = $\log a/\log b$.
Then one has to design a geometrical iterator such that it would take $a$ steps
to advance a distance equal to $b$.

This problem can be solved automatically
by means of grammatical evolution. Our genetic algorithm
acts on genotypes consisting of vectors of integers and
makes use of a fixed grammar to translate the genotypes
into an intermediate level, which can be interpreted as a single
rule for an L system which, together with a turtle graphic
interpretation, generates the final phenotype: a fractal
curve with a dimension as approximate as desired to the desired value.
The algorithm can be described as follows:

\begin{enumerate}
\item Generate a random population of 64 vectors of eight
integers in the $[0, 10]$ or the $[0, 255]$ interval 
(the latter case introduces genetic code degeneracy).
All the genotypes in the initial population have the same length.
Subsequent populations may contain individuals with genotypes of different lengths.
\item Translate every individual genotype into a word in the
alphabet $F, +, -$ as indicated below.
\item Using the algorithm described in section 3, compute the dimension of the fractal curve
represented by the D0L system which uses the preceding word as a generator.
\item Compute the fitness of every genotype as
$(\text{target}-\text{dimension})^{-1}$.
\item Order the 64 genotypes from higher to lower fitness.
\item If the highest-fitness genotype has a fitness higher than
the target fitness, stop and return its phenotype.
\item From the ordered list of 64 genotypes created in step 5,
remove the 16 genotypes with least fitness (leaving 48)
and take the 16 genotypes with most fitness. Pair these
16 genotypes randomly to make eight pairs. Each
pair generates another pair, a copy of their parents,
modified according to four genetic operations (see below). The new
16 genotypes are added to the remaining population of
48 to make again 64, and their fitness is computed as in
steps 2 to 4.
\item Go to step 5.
\end{enumerate}

The algorithm has three input parameters: the target
dimension (used in step 4), the target minimum fitness (used in step 6) and the angle step
for the turtle graphics interpretation (used in step 3).

In step 2, the following grammar is used to translate the genotype of one individual 
into its equivalent intermediate form (the generator for an L system representing
a fractal curve):

$0: F::=F$

$1: F::=FF$

$2: F::=F+$

$3: F::=F-$

$4: F::=+F$

$5: F::=-F$

$6: F::=F+F$

$7: F::=F-F$

$8: F::=+$

$9: F::=-$

$10: F::=\lambda$

The translation is performed according to the following developmental algorithm:

\begin{enumerate}
\item The axiom (the start word) of the grammar is assumed to be $F$.
\item As many elements from the remainder of the genotype
are taken (and removed) from the left of the genotype
as the number of times the letter $F$ appears in the current word. If there remain
too few elements in the genotype, the required number is completed circularly.
\item Each $F$ in the current word is replaced by the right-hand
side of the rule with the same number as the integers
obtained by the preceding step. With genetic code degeneracy, the remainder of each integer modulo
11 is used instead. In any derivation, the trivial rules $+::=+$ and $-::=-$ are also applied.
\item If the genotype is empty, the algorithm stops and returns the last derived word.
\item If the derived word does not contain a letter $F$, the whole word is
replaced by the axiom.
\item Go to step 2.
\end{enumerate}

The four genetic operations mentioned in step 7 of the genetic algorithm are the following:
\begin{itemize}
\item \emph{Recombination} (applied to all the generated genotypes).
Given a pair of genotypes, $(x_1, x_2, ..., x_n)$ and
$(y_1, y_2, ..., y_m)$, a random integer is generated in the
interval $[0, \min(n, m)]$. Let it be $i$. The resulting
recombined genotypes are $(x_1, x_2, ..., x_{i-1}, y_i, y_{i+1} ,
..., y_m)$ and\\$(y_1, y_2, ..., y_{i-1}, x_i, x_{i+1}, ..., x_n)$.
\item \emph{Mutation}, applied to $n_1$ per cent of the generated genotypes if
both
parents are equal, to $n_2$ per cent if they are different. It
consists of replacing a random element of the genotype vector
by a random integer in the same interval.
\item \emph{Fusion}, applied to $n_3$ per cent of the generated genotypes. The
genotype is replaced by a catenation of itself with a
piece randomly broken from either itself or its brother's
genotype. (In some tests, the whole genotype was used,
rather than a piece of it.)
\item \emph{Elision}, applied to 5 per cent of the generated genotypes. One
integer in a random position of the vector is eliminated.
The last two operations make it possible to generate longer or shorter
genotypes from the original eight element vectors. 	
\end{itemize}

%% file: cap3.tex
\section{Heavy tail distributions}
\label{heavytail}

Heavy tail distributions are probabilistic distributions which exhibit an
asymptotic hyperbolic decrease, usually represented as
\begin{equation}
\Pr\{|X|>x\}\sim Cx^{-\alpha},
\label{typicaltail}
\end{equation}
where $\alpha$ is a positive constant. Distributions with
this property  have been used to model ramdom variables whose extreme values are
observed with a relatively high probability.

Work on these probability distributions can be traced to
Pareto's \citeyear{Pareto65}
work on the earning distribution 
or to Levy's \citeyear{Levy37} work on the properties of stable
distributions. A fundamental advance in the use of heavy tail
distributions for  was provided by Mandelbrot's work
\citeyear{Mandelbrot60,Mandelbrot63}
on the application of fractal behavior and self-similarity to the modeling of real-world phenomena,
which he used to introduce stable distributions to model price changes in the stock exchange.
Heavy tail distributions have also been used in areas such as
statistical physics, wheather prediction, earthquake prediction,
econometrics and risk theory \cite{Embrechts97,Mandelbrot83}.
In more recent times, these distributions have been used to model waiting
times in the World Wide Web \cite{Willinger94} or the computational cost of
random algorithms \cite{Gomes03,Gomes00,Gomes1998,Gomes97}.

% Different procedures have been proposed to estimate the parameters
% of Pareto and stable distributions. A comparative revision of those methods
%can
% be found in \cite{Hughey91}.  \citeasnoun{Adler98} offer another more recent
% and
% detailed revision.

For many purposes, the only relevant parameter of a heavy tail distribution is its
\emph{characteristic exponent} $\alpha$, which determines the ratio of
decrease of the tail and the probability of occurrence of extreme events. In
this work we only consider  heavy tail distributions where $\alpha$
belongs to the $(0,2)$ interval, with positive support ($\Pr\{0 \leq X
< \infty\}=1$).

The existence or inexistence of the different moments of a distribution is fully determined by the
behavior of its tail: $\alpha$ can also be regarded as the exponent of the
maximum finite moment of the distribution, in the sense that moments of $X$ of order less than
$\alpha$ are finite, while moments of order equal or greater are infinite. For instance, when $\alpha
= 1.5$, the distribution has a finite average and an infinite variance, while for
$\alpha=0.6$ both average and variance are infinite.

% The first is the maximum likelihood estimator introduced by
% \citeasnoun{Hill75} and studied by Hall \citeasnoun{Hall82} (thus called the
% Hill-Hall estimator) is  a parametric estimator applicable to observations
% generated by a Pareto distribution, although once truncated for extreme
% observations provides an asymptotically unbiased estimator for the behavior of
% the tail.
%The Hill-Hall estimator is the most natural and used of all the estimators.
% 

% Simulations in \cite{Crato00} proved that these
% estimators perform successfully with the upper truncation, while the classic
% Hill-Hall estimator becomes seriously biased.

\subsection{Estimation of the characteristic exponent}

Many procedures have been used to estimate $\alpha$
\cite{Hughey91,Adler98,Crato00}. Two of them have
received the most extensive usage. The first uses a maximum likelihood
estimator, the second applies a simple regression method.

An important issue while estimating $\alpha$ is how to tackle censored
observations when extreme data are not available. Consider, for instance,
physical phenomena such as wind velocity or earthquake magnitude, where
heavy tail distributions have been considered appropriate.
In these cases, extreme measures are non-observable,
since very strong hurricanes or highly destructive earthquakes will damage the
measuring instruments. In the process of financial data, such as
stock exchange rates, heavy tail models have also been used
\cite{Lima97}. In moments of high volatility, when extreme data usually
appear, many stock exchange markets introduce rules to limit transactions or
even close the market, to prevent them from taking place. Consider finally the case of
random algorithms: the computational costs of some problems are so high,
that the algorithms have no alternative but to interrupt the execution
and start again with a different random seed. In those cases, computational
costs are not observable beyond a certain threshold \cite{Gomes98}. Thus the
censorship of extreme values needs to be considered by available estimators. 

% We will only consider the asymptotic behavior of the distribution tails and an
% estimation of the $\alpha$ parameter.

Let $X_{n1} \leq X_{n2} \leq \ldots X_{nn}$ be the ordered statistics, 
i.e. the ordered values in the sample
$X_{1},X_{2},\ldots,X_{n}$. Let $r<n$ be the truncation value which separates
normal from extreme observations.

The adapted Hill-Hall estimator for censored observations is:
\begin{equation}
  \hat{\alpha}_{r,u}=(\frac{1}{r}\sum_{j=1}^{r-1}\ln X_{n,n-r+j} +
  \frac{u+1}{r}\ln X_{n,n}-\frac{u+r}{r}\ln X_{n,n-r})^{-1}.
\label{adapted Hill-Hall estimator}
\end{equation}
% The variance of this estimator is:
% getting:
% \begin{equation*}
% \widehat{\text{Var}}(\hat{\alpha})=\frac{\alpha^2(r+1)^2}{r^2(r-1)}.
% \end{equation*}
In this notation, $n$ is the number of observed data, $r+1$ is the number
of larger observations selected and  $u$ is the number of non-observed extreme
values.
If all the data are observable, $u=0$ and equation (\ref{adapted Hill-Hall
estimator}) becomes the classic Hill-Hall estimator.

In heavy tail distributions, the ratio of decay of the estimated
density is hyperbolic (slower than a exponential decay). Thus the
one-complement of the accumulated distribution function,
$\overline{F}(\cdotp)$, also shows a hyperbolic decay.
\begin{equation}
  \overline{F}(x)=1-F(x)=\Pr\{X>x\}\sim Cx^{-\alpha}.
\label{base maxima verosimilitud}
\end{equation}
Therefore, for a heavy-tail variable, a log-log graph of the frequency
of observed values larger than $x$ should show an approximately linear decay 
in the tail. Moreover, the slope of the linear decaying graph is in itself
an estimation of $\alpha$. This can be contrasted with a exponentially decaying
tail, where a log-log graph shows a faster-than-linear tail decay.

This simple property, besides giving visual evidence of the presence of a heavy
tail, also gives place to a natural regression estimator based on
equation \ref{base maxima
verosimilitud}, the least-squares estimator \cite{Adler98}, which can be
expressed in terms of a selected number of
extreme observations. 
Assume that we have a sample of $k=n+u$ independent identically distributed
random variables. Assume also that we only observe the $n$ smallest values of random
variable $X$ and therefore have the ordered statistics $X_{n1} \leq
X_{n2} \leq \ldots \leq X_{nn}$. Assume that, for $X_{n,n-r} \leq X \leq X_{nn}$,
the tail distribution has a hyperbolic decay. The 
least-square regression estimator for the $\alpha$ exponent is
\begin{equation}
\hat{\alpha}=-\frac{\sum l_{i}\log X_{ni} - \sum l_{i} \sum \log
X_{ni}/(r+1)}{\sum(\log X_{ni})^2 - \sum(\log X_{ni})^2/(r+1)},
\label{regresion}
\end{equation}
where $l_{i}=\log \frac{n+u-i}{n+u}$ and the sums go from $i=n-r$ to
$i=n$. If all the values in the sample $k=n+u$ can be observed,  then $u=0$ and
$k=n$.

\subsection{Leptokurtic distributions vs. heavy tail distributions}
The name heavy tail, applied to a class of distributions,
expresses their main property: the large proportion of total probability
mass concentrated in the tail, which reflects its (hyperbolic) slow decay
and is the reason why all the moments of a heavy tail distribution are infinite,
starting at a given order.

% \begin{figure}[t]
% \centering
%  \includegraphics[width=\textwidth]{normalcauchylevy}
% \vspace*{-6mm}
% \caption{normal, Cauchy and Levy density functions.}
% \label{normalcauchylevy}
% \end{figure}
% 
% \begin{table}[b]
% \centering
% \begin{tabular}{|c|l|l|l|}
% \hline c & Normal & Cauchy & L�y \\ \hline 0 & 0.5000 & 0.5000 &
% 1.0000 \\ \hline 1 & 0.1587 & 0.2500 & 0.6837 \\ \hline 2 & 0.0228 &
% 0.1476 & 0.5205 \\ \hline 3 & 0.001347 & 0.1024 & 0.4363 \\ \hline 4 &
% 0.00003167 & 0.0780 & 0.3829 \\ \hline 5 & 0.0000002866 & 0.0628 &
% 0.3453 \\ \hline
% \end{tabular}
% \caption{A comparison of the tail probability, $\Pr\{X>c\}$
% for the standard normal, Cauchy and L\'evy distributions.}
% \label{tablanormalcauchylevy}
% \end{table}
% 
% Figure \ref{normalcauchylevy} compares three distributions: standard normal,
% Cauchy and L\'evy. The key property to be observed is the dramatic difference
% in the decay of the tails of these three distributions. Table
% \ref{tablanormalcauchylevy} shows the total
% proportion of the mass probability in the tail of these distributions for
% several values.
% It can be observed that this quantity vanishes quickly for the standard normal
% distribution,
% while the other two contain a significant probability mass in their tails.
% The lower the value of $\alpha$ (the stability index of the distribution), the
% heavier is the tail.
% For instance, Cauchy's distribution has $\alpha = 1.0$, while L\'evy's
% distribution has
% $\alpha = 0.5$.

The concept of \emph{kurtosis} is also related to the tail heaviness. The
kurtosis of a
distribution is the amount $\mu_{4}/\mu_{2}^{2}$, where $ \mu_{2}$ and
$\mu_{4}$ are the second and fourth centralized moments ($\mu_{2}$ is the
variance). The kurtosis is independent of the localization and scale parameters
of a distribution. Kurtosis is high, in general, for a distribution with a high
central peak and long tails.

\begin{figure}[h]
\centering \includegraphics[width=\textwidth]{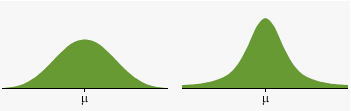}
\caption{Low kurtosis vs. high kurtosis. The probability density function
on the right has a higher kurtosis than the left:
its center part has a higher peak and its tails are heavier.}
\label{kurtosis}
\end{figure}

The kurtosis of the standard normal distribution is 3. A distribution with
a kurtosis higher than 3 is called \emph{leptokurtic} as opposite to
\emph{platokurtic} (see fig. \ref{kurtosis}) . In a similar way to
heavy tail distributions, a leptokurtic distribution has long tails with a
considerable concentration of probability. However, the tail of a leptokurtic
distribution decays faster than that of a heavy tail distribution: all
the moments in a leptokurtic distribution can be finite, in a strong contrast
with a heavy tail distribution where, at most, the first two moments are finite.

%% file: cap4-2.tex
\section{Heavy tails in Grammatical Evolution}

Randomized algorithms with a high execution time variability
are suspetc of hiding a heavy tail distribution. In the present section
we provide empirical evidence that our GE algorithm for the automatic
generation of fractal curves may exhibit a heavy tail behavior
which can be exploited to improve the performance.

\begin{table}[h]
\centering
\begin{tabular}{|p{2cm}|l|l|p{4cm}|}
\hline dimension & angle (degrees) & \# experiments & \#
generations range\\ 
\hline 1.1   & 45 & 10 & [37, 9068]\\ 
\hline 1.1   & 60 & 4  & [119, 72122]\\ 
\hline 1.2   & 45 & 8  & [188, 11173]\\ 
\hline 1.2   & 60 & 10 & [21, 750]\\ 
\hline 1.3   & 45 & 9  & [50, 18627]\\ 
\hline 1.3   & 60 & 4  & [14643, 66274]\\ 
\hline 1.25  & 60 & 2  & [1198, 3713]\\ 
\hline 1.255 & 60 & 15 & [1, 2422]\\ 
\hline 1.2618595... & 60 & 4 & [1, 2]\\
\hline 1.4 & 45 & 10 & [79, 781]\\ 
\hline 1.4 & 60 & 10 & [33, 1912]\\
\hline 1.5 & 45 & 11 & [52, 11138]\\ 
\hline 1.5 & 60 & 8 & [12, 700]\\
\hline 1.6 & 45 & 5 & [275, 3944]\\ 
\hline 1.6 & 60 & 1 & [116, 913]\\
\hline 1.7 & 45 & 2 & [585, 1456]\\ 
\hline 1.7 & 60 & 8 & [18, 1221]\\
\hline 1.8 & 45 & 2 & [855, 2378]\\ 
\hline 1.8 & 60 & 13 & [69, 3659]\\
% \hline 1.9 & 72 & 1 & 5467\\ 
% \hline 1.95 & 90 & 1 & 956\\ 
% \hline 2 & 45 & 5 & 1\\ 
% \hline 2 & 90 & 5 & 1\\ 
\hline
\end{tabular}
\caption{Number of generations needed to generate a
fractal curve with a given dimension in a set of experiments.}
\label{tabla-ortega03}
\end{table}

\citeasnoun{Ortega03} provides data about
different executions of the same algorithm to generate fractal curves with the same dimensions,
using different random seeds. The numbers of generations needed to reach the target
differ in up to two orders of magnitude (see table \ref{tabla-ortega03}).

% This means that we have signaled a \emph{purely computational}
% approach to take advantage of the existence of heavy tail distributions.

% Our goal, as presented in Section \ref{} is to generate a Lindenmayer grammar
% which
% represents a fractal curve with a dimension approximately equal to a required 
% \emph{target}.
% The following fitness criterion is used:
% \begin{equation}
% \text{fitness}(x)=\frac{1}{|\text{target}-\text{dimension}(x)|},
% \end{equation}
% where $\text{dimension}(x)$ is a function defined in the following way:
% \begin{center}
% {\footnotesize
% \begin{tabular}{ccccc}
% binary genotype & & Lindenmayer grammar & & fractal dimension \\
% $x$ & $\longrightarrow$ & $L_{x}$ & $\longrightarrow$ & $d$ \\
% \end{tabular}
% }
% \end{center}
% i.e., by means of a genotype-phenotype translation process,
% a variable length binary genotype is converted into a Lindenmayer grammar,
% from which the dimension of the fractal curve it represents is computed
% using the algorithm detailed in \cite{Alfonseca00,Alfonseca01}. Once the
% dimension of the fractal curve represented by
% the grammar is sufficiently near to the \emph{target}, a graphic
% interpretation
% of the Lindenmayer grammar (either vector or turtle graphics) generates
% the required fractal curve.

\begin{figure}[h]
\centering 
\includegraphics[width=\textwidth]{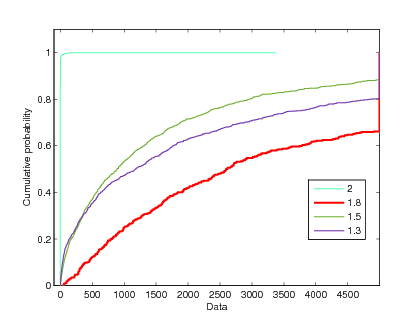}
\caption{Empirical distribution function of the number of generations needed to
reach a solution for several fractal dimensions: 1.3, 1.5, 1.8 and 2.}
\label{cdf}
\end{figure}

Figure \ref{cdf} shows the empirical distribution of the number of
generations needed to find a solution, i.e. 
\begin{equation} 
F(x)=\Pr\{\text{number of generations to reach a solution}\leq x\}
\end{equation}
for four different fractal dimensions: 1.3, 1.5, 1.8 and 2. 
%The fractal dimensions generated by this algorithm belongs to the $(1,2]$
% interval. 
The empirical distribution functions
have been obtained by running 1,000 executions with 1,000 different
independent random seeds.
At the end of each execution, the number of generations needed to
reach a solution is recorded.
We took a \emph{censorship} value equal to $\tau = 5,000$ generations,
meaning that, if an execution needs over 5,000 generations, it is stopped and
marked as non-observable.

\begin{table}[h]
\centering
\begin{tabular}{|r|r|r|r|r|}
\hline dimension & 1.3 & 1.5 & 1.8 & 2 \\ \hline observable & 80.5\%
& 88.3\% & 66.2 \% & 100\% \\ \hline non-observable & 19.5\% & 12.7\% &
38.2 \% & 0\% \\ \hline
\end{tabular}
\caption{Percentages of observable and non-observable executions for a
censorship value
$\tau = 5,000$ generations.}
\label{observables}
\end{table}

Table \ref{observables} shows the percentages of non-observable executions
in our experiments. This percentage is quite high, specially for
dimensions 1.8 and 1.3. The empirical distribution functions may be used
to test whether the distribution has a heavy tail.

In the previous section (definition \ref{typicaltail}) we saw that a random variable has
a heavy tail behavior if it shows an \emph{asymptotic} hyperbolic decay,
although that behavior can also be shown in its whole support.
In the figures displayed in this section, only the extreme values are shown,
therefore we had to choose a parameter $r$ to truncate the non-extreme
observations. Usually $r$ takes values in the $[1\%,25\%]$ interval; 
we will use the set \{1\%,
2.5\%, 5\%, 10\%, 15\%, 20\%\}, as recommended by  \citeasnoun{Crato00}.

\begin{figure}[ht]
\centering
\begin{tabular}{cc}
\includegraphics[width=170pt]{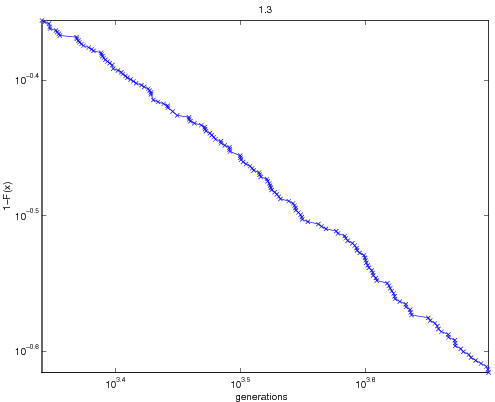} &
\includegraphics[width=170pt]{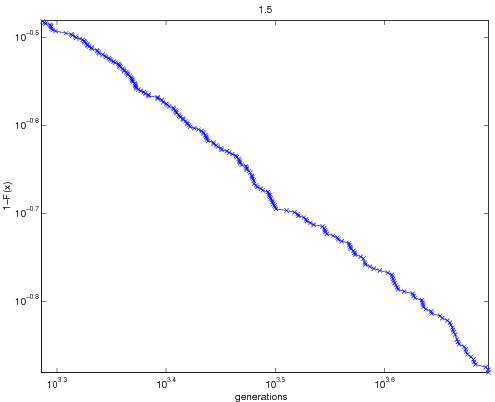} \\
\includegraphics[width=170pt]{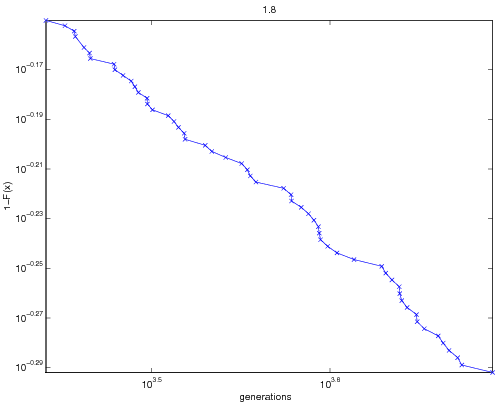} &
\includegraphics[width=170pt]{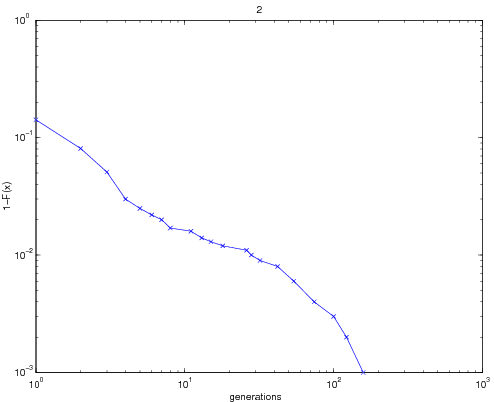} \\
\end{tabular}
\caption{Log-log graph of the tail of ($r$=20\%) distributions
for dimensions 1.3, 1.5, 1.8 and 2.}
\label{ca1cdf}
\end{figure}

Figure \ref{ca1cdf} shows the log-log graphs of the distribution tails for
fractal dimensions 1.3, 1.5, 1.8 and 2. Notice the linear decay of function
$\log \overline{F}(x)$, in contrast with exponential decay distributions,
where the decay of $\log \overline{F}(x)$ is faster than linear.

\begin{figure}[h]
\centering
 \includegraphics[width=\textwidth]{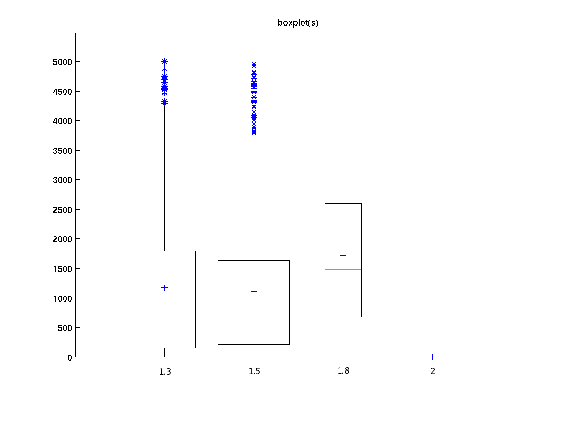}
\vspace*{-0.5cm}
\caption{\emph{Box-and-whisker} type graphs for dimensions
1.3, 1.5, 1.8 and 2.}
\label{boxplots}
\end{figure}

The averages for dimensions
1.3, 1.5 and 1.8 are $E(X_{1.3})=1,173$,
$E(X_{1.5})=1,108$ and $E(X_{1.8})=1,721$.
It can be seen that, with a number of generations almost 5 times above
their averages, respectively over 10\%, 20\% and 30\% executions
have not finished.

Figure \ref{boxplots} displays four
\emph{box-and-whisker} graphs, which give rise to three remarkable conclusions:
\begin{itemize}
\item The median (the dashed line within the box in fig. \ref{boxplots}) is much
smaller than the average (the cross `+' within the box) for dimensions 1.3, 1.5
and 1.8. This suggests that the average of these distributions is biased by the
size of the sample, which means that they may have an infinite asymptotic
average typical of heavy tail distributions.
\item The sample distribution is strongly biased towards high execution times, 
indicating a right-hand-side heavy tail. This can be seen in the fact 
that the lower interquartilic distance (the difference between the first 
quartile - the lower segment of the box - and the the median - the green 
line) is shorter than the upper interquartilic distance (the difference 
between the median and the third quartile - the upper segment of the box). 
Besides this, the distance between the minimum and the first quartile 
is much less than the distance between the maximum (the highest 
point of the graph) and the third quartile.
\end{itemize}

\subsection{Estimating the characteristic exponent}

The preceding section provides visual evidence for a heavy tail behavior in
dimensions 1.3, 1.5, 1.8. Evidence for this behavior is weaker in dimension 2,
but also present in, for instance, the linear decay observed in figure
\ref{ca1cdf}. In this section we estimate the
characteristic exponent for these distributions, using the estimators presented
in section \ref{heavytail}.

First we compute the Hill-Hall
estimator adapted for censored observations, (equation \ref{adapted Hill-Hall
estimator}).
%As already mentioned, we shall use the set $r$=\{1\%, 2.5\%, 5\%, 10\%, 15\%,
%20\%\}.
\begin{table}[h]
\centering
\begin{tabular}{|c|c|c|c|c|c|c|}
\hline \multirow{2}{1.8cm}{dimension} & \multicolumn{6}{c|}{$r$} \\
\cline{2-7}

    & 1\% & 2.5\% & 5\% & 10\% & 15\% & 20\% \\ \hline 1.3 & 0.7827 &
0.6796 & 0.8312 & 0.7953 & 0.7634 & 0.7084 \\ \hline 1.5 & 1.1765 &
1.2400 & 1.0952 & 1.0595 & 0.9952 & 0.9418 \\ \hline 1.8 & 0.3649 &
0.4855 & 0.6746 & 0.5759 & 0.5657 & 0.5705 \\ \hline 2 & 0.7656 &
0.6043 & 1.0403 & 1.0463 & 0.7732 & 1.0309\\ \hline
\end{tabular}
\caption{Estimations of $\alpha$ for dimensions 1.3, 1.5, 1.8 and
2 using the adapted Hill-Hall estimator.}
\label{estimaciones hill-hall} 
\end{table}

%Table \ref{estimaciones hill-hall} shows the estimations of
%$\alpha$ obtained using the Hill-Hall estimator adapted to censored
%observations.

Table \ref{estimaciones hill-hall} confirms that these distributions are heavy
tailed, since all the values in
the table are less than 2, the limit for heavy tail distributions.

For dimension 1.3, all the estimations (for all values of
$r$) are less than 1, which means that this distribution does not have
neither a finite average nor a finite variance. The same happens for dimension 1.8
even in a stronger way, as the values of $\alpha$ are even smaller
(all are below 0.7).

Dimensions 1.5 and 2 provide examples of heavy tail distributions
with a characteristic exponent $\alpha$ between 1 and 2.
These distributions have a finite average, but an infinite variance,
indicating that their right heavy tail is lighter than in the other two
distributions.

\begin{figure}[h]
\centering
\begin{tabular}{cc}
\includegraphics[width=170pt]{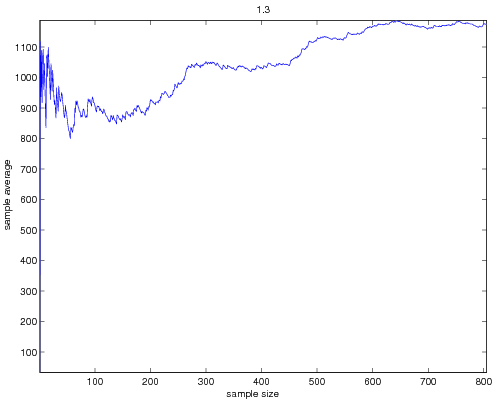} &
\includegraphics[width=170pt]{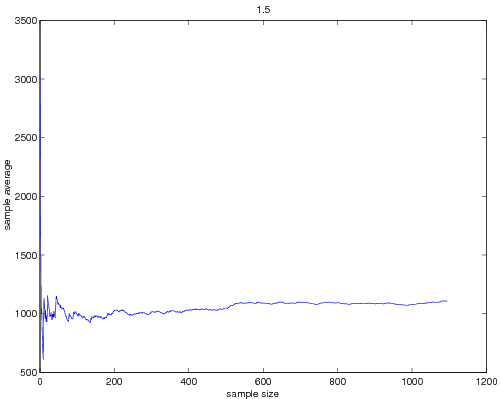} \\
\includegraphics[width=170pt]{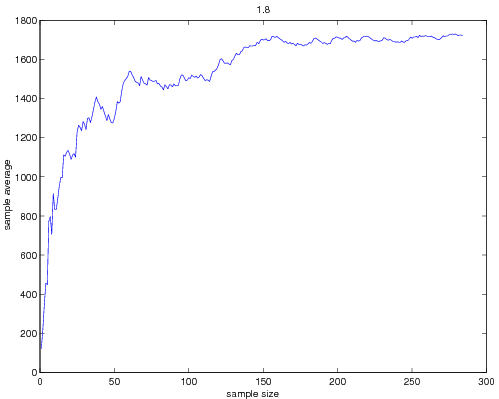} &
\includegraphics[width=170pt]{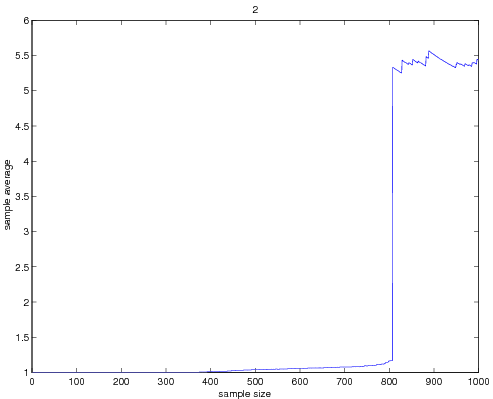}
\end{tabular}
\caption{Evolution of the sample average as a function of the sample size
for dimensions 1.3, 1.5, 1.8 and 2.}
\label{medias incrementales}
\end{figure}

Figure \ref{medias incrementales} displays the erratic behavior of the sample
average as a function of the sample size.

\begin{table}[h]
\centering
\begin{tabular}{|c|c|c|c|c|c|c|}
\hline \multirow{2}{1.8cm}{dimension} & \multicolumn{6}{c|}{$r$} \\
\cline{2-7} & 1\% & 2.5\% & 5\% & 10\% & 15\% & 20\% \\ \hline 1.3 &
0.6952 & 0.7528 & 0.7715 & 0.7904 & 0.7692 & 0.7345 \\ \hline 1.5 &
1.1318 & 1.3790 & 1.0886 & 0.9664 & 0.9786 & 0.9721 \\ \hline 1.8 &
0.3310 & 0.5220 & 0.7285 & 0.6424 & 0.5762 & 0.5762 \\ \hline 2 &
$\approx$0 & $\approx$0 & 0.2554 & 0.4821 & 0.6008 & 0.6667 \\ \hline
\end{tabular}
\caption{Estimations of $\alpha$ for fractal dimensions 1.3, 1.5, 1.8 and
2, using the regression estimator.}
\label{estimaciones regresion} 
\end{table}

To verify the reliability of our characteristic exponent estimation,
table \ref{estimaciones regresion} shows the estimations obtained
using the regression estimator described in a previous section (equation
\ref{regresion}),
which is  considered slightly less robust than the maximum likelihood
estimator (adapted Hill-Hall). The results of this estimator can be seen to be
consistent with those of the adapted Hill-Hall estimator.

\subsection{Tail truncation}

As mentioned before, in practice one has to select the GE maximum number of
generations for specially difficult problems. In other words, an appropriate censorship value $\tau$
must be chosen, so that the algorithm does not become stagnated in the extreme values of
the distribution tail. As a consequence, the tail is truncated.
The selection of the value of $\tau$ depends on the problem and
the algorithm. Ideally, only a small portion of tail should be truncated,
but  this may be prohibitive from the computational point of view.

If the truncation is set at a small number of generations, it will be harder to distinguish
between heavy tail and leptokurtic distributions. From a practical point of view, this
is not a problem, if there are strong indications that the tail exhibits at least one of the two
behaviors. A heavy tail behavior is not a necessary condition to accelerate
randomized search methods. In fact, it has been proved that the efficiency of the search
in leptokurtic distributions can be improved by randomized backtracking
\cite{Gomes03}.
However, with a heavy tail distribution, the occurrence of long executions will be
more frequent than with a leptokurtic distribution,
making it possible to obtain a higher potential acceleration.

\begin{figure}[h]
\centering
\begin{tabular}{cc}
\includegraphics[width=170pt]{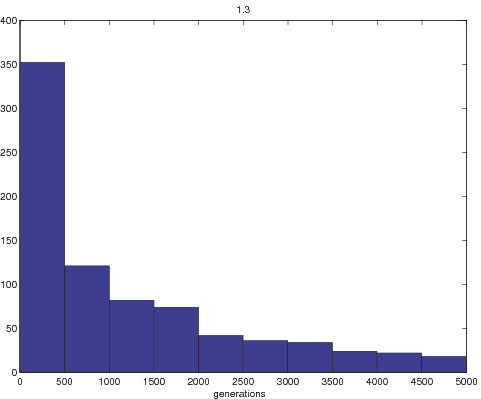} &
\includegraphics[width=170pt]{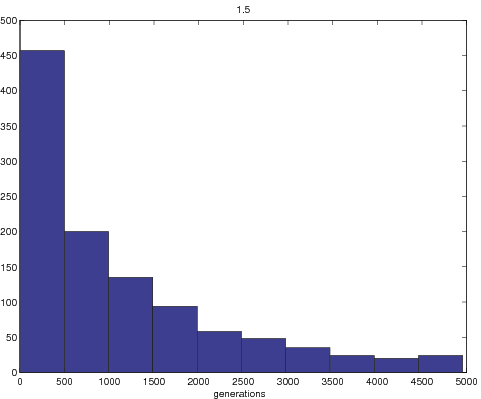} \\
\includegraphics[width=170pt]{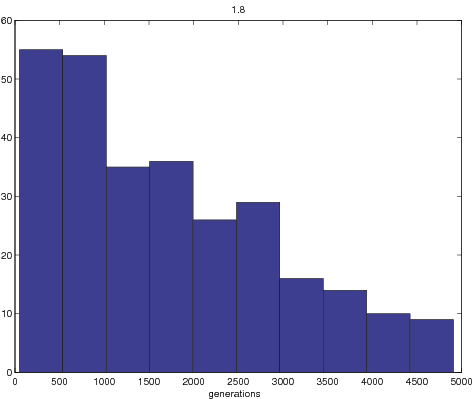} &
\includegraphics[width=170pt]{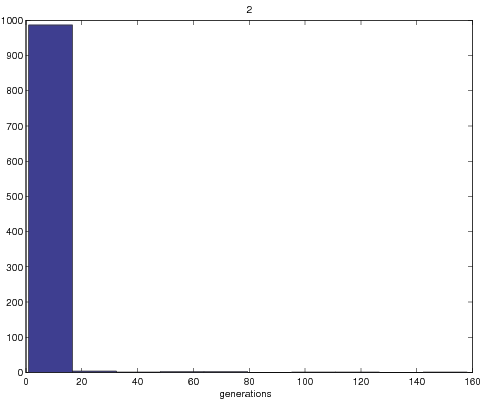}
\end{tabular}
\caption{Histograms with the execution samples obtained for fractal dimensions
1.3, 1.5, 1.8 and 2.}
\label{histogramas}
\end{figure}

\begin{table}[h]
\centering
\begin{tabular}{|c|r|r|r|r|}
\hline dimension & 1.3 & 1.5 & 1.8 & 2\\ \hline $\mu_{2}$ & 1.6171e+06
& 1.3532e+06 & 1.5496e+06 & 69.9039\\ \hline $\mu_{4}$ & 9.0640e+12 &
7.6389e+12 & 6.0015e+12 & 9.6330e+05 \\ \hline $\text{kurt}(x)$ &
3.4575 & 4.1642 & 2.4817 & 197.1335\\ \hline
\end{tabular}
\caption{Kurtosis computation for dimensions 1.3, 1.5, 1.8 and 2.}
\label{tabla kurtosis}
\end{table}
Table \ref{tabla kurtosis} shows the kurtosis for the 4 fractal dimensions considered.
Remember that, if this value is greater than 3 (the kurtosis for a normal
distribution) the distribution is leptokurtic (with abrupt peaks and heavy tails),
otherwise it is platokurtic (with smooth peaks and light tails).
In our case, fractal dimensions 1.3, 1.5 and 2 are seen to be leptokurtic, while
dimension 1.8 is platokurtic.
 Figure \ref{histogramas} shows the histograms built for the execution samples for
dimensions 1.3, 1.5, 1.8 and 2.

This section ends  with the conclusion that there exists an application of GE,
automatic fractal generation, whose distributions exhibit a heavy tail behavior,
besides being leptokurtic in many cases. In the next section we show that it is possible
to take advantage of this probabilistic characterization to increase the performance
of GE and yield a fast fractal generation algorithm.

%% file: cap5.tex
\section{Restart strategies}

We have shown that our algorithm may give rise to computational efforts with a leptokurtic or heavy tail
distribution. This may be due to the fact that the algorithm makes \emph{bad
choices} more frequently than expected, leading the search to a dead-end in the search space, where no
solution of the required fitness exists.

The algorithm seems to be more efficient at the beginning of the
search, which suggests that a sequence of short executions, compared to a single
long execution, may give rise to a better use of the computational resources.
In this section we show that the algorithm may be accelerated by the use of
several \emph{restart strategies}.

\subsection{Restarts with a fixed threshold}

\begin{figure}[h]
\centering 
\includegraphics[width=\textwidth]{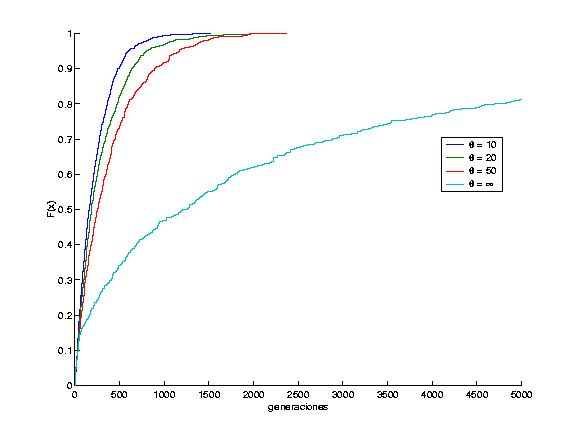}
\caption{Function $F(x)$ for several values of the restart threshold
$\theta \in \{10, 20, 50, \infty\}$ applied to fractal dimension 1.3.}
\label{cdfrecomienzos-1.3}
\end{figure}

Figure \ref{cdfrecomienzos-1.3} displays the result of a \emph{restart strategy with a fixed threshold}
applied to the generation of a fractal curve with dimension 1.3. This is the
simplest strategy:
once the algorithm has been working for a predefined number of generations $\theta$,
without reaching the desired goal, a new execution is started with a different
random seed. As the figure shows, the failure rate after 500
generations is 70\% ($F(500)=0.3$), while this percentage falls to 10\% using
restarts with a threshold $\theta = 10$ generations.

Such an improvement is typical of heavy tail distributions. The fact that the experimental curve
has been so dramatically moved towards the beginning of the support is a clear
indication that the heavy tail character of the original distribution has
disappeared in the modified algorithm.

\begin{figure}[h]
\centering
 \includegraphics[width=\textwidth]{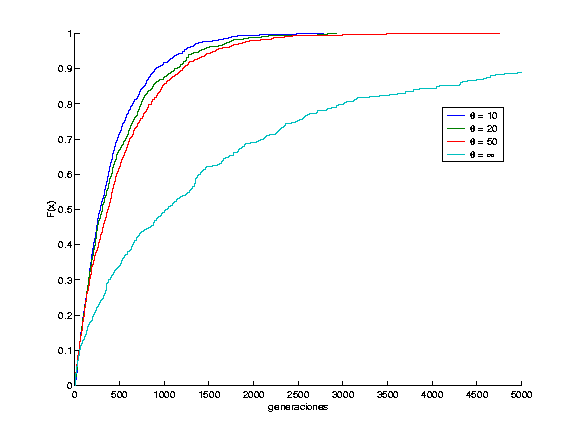}
\caption{Function $F(x)$ for several values of the restart threshold
$\theta \in \{10, 20, 50, \infty\}$ applied to fractal dimension 1.5.}
\label{cdfrecomienzos-1.5}
\end{figure}

\begin{figure}[h]
\centering 
\includegraphics[width=\textwidth]{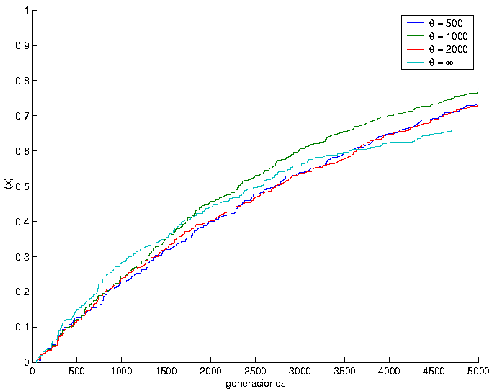}
\caption{Function $F(x)$ for several values of the restart threshold
$\theta \in \{500, 1000, 2000, \infty\}$ applied to fractal dimension 1.8.}
\label{cdfrecomienzos-1.8}
\end{figure}

\begin{figure}[h]
\centering \includegraphics[width=\textwidth]{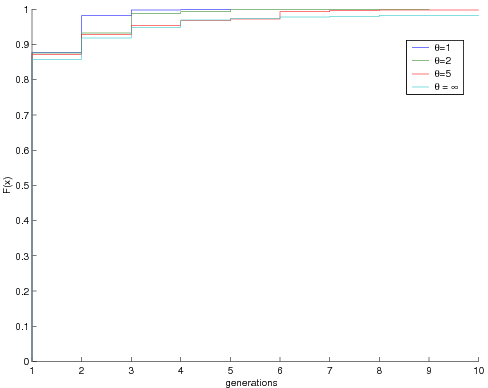}
\caption{Function $F(x)$ for several values of the restart threshold
$\theta \in \{1, 2, 5, \infty\}$ applied to fractal dimension 2.}
\label{cdfrecomienzos-2}
\end{figure}

Figures \ref{cdfrecomienzos-1.5}, \ref{cdfrecomienzos-1.8} and
\ref{cdfrecomienzos-2}
clearly show that the restarts make the tail of the distributions
\emph{lighter}, thus providing
a mechanism to handle heavy tail and leptokurtic distributions. 

\begin{table}
\centering
\begin{tabular}{|r|r|r|}
\hline $\theta$ & \% solved & average cost \\ \hline 2 & 100\% &
382.6740 \\ \hline 4 & 100\% & 277.5730 \\ \hline 8 & 100\% & 207.8240
\\ \hline 16 & 100\% & 271.3980 \\ \hline 32 & 100\% & 345.2680 \\
\hline 64 & 100\% & 407.2460 \\ \hline 128 & 100\% & 621.1770 \\
\hline 256 & 99.8\% & 830.4220 \\ \hline 512 & 98.5\% & 985 \\ \hline
1024 & 96.4\% & 1,367 \\ \hline 2048 & 93.7\% & 1,909 \\ \hline
\end{tabular}
\caption{Percentage solved and  average cost for several
threshold values in the fractal dimension 1.3 experiment.}
\label{varios umbrales}
\end{table}

\begin{figure}
\centering 
\includegraphics[width=\textwidth]{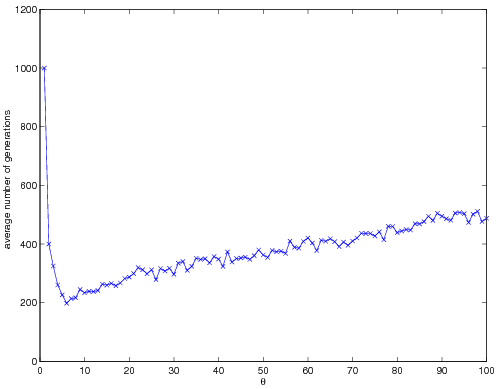}
\caption{The effect of restarts with fixed $\theta$ on the solution costs for fractal dimension 1.3.}
\label{umbrales-1.3}
\end{figure}

Different fixed thresholds give rise to different average times needed to reach
a solution.
Table \ref{varios umbrales} and figure \ref{umbrales-1.3} show that the
threshold
value $\theta = 6$ minimizes the expected cost for fractal dimension 1.3,
making it the optimal threshold $\theta^{*}$.
For threshold values larger than the optimal, the heavy tail behavior at the right
of the median dominates the average cost, while below the optimal value the success
percentage is too small and too many restarts are required. Anywhere,
many non-optimal choices provide a considerable acceleration of the algorithm.

It has been proven that the use of a fixed restart threshold
$\theta$ with a heavy tail distribution eliminates this behavior in such a way
that all the moments of the new distribution become finite \cite{Gomes00}.

% Walsh \cite{Walsh99} introduced a new restart strategy, inspired by
% \citeasnoun{Luby93} analysis,
% where the value of the threshold is increased geometrically after every
%restart.
% This strategy is less sensible to the details of the underlying distribution. 

% In
% practice, many SAT solvers use a default threshold value, which is
% incremented linearly after a given number of restarts, thus guarantying that
%the  solver
% is complete at the limit \cite{Moskewicz01}.

\subsection{Restart sequences}

The idea of a fixed threshold comes from theoretical results by 
\citeasnoun{Luby93}, which describe optimal restart policies.
It can be proven that if the time distribution of the execution is completely
known and therefore $\theta^{*}$ can be calculated a priori, 
restarting every $\theta^{*}$  generations yields the minimum average execution
time. 

% In our case, therefore, the \emph{optimal} strategy
% is simply a sequence of executions with a fixed threshold, since we assume
% that
% the empirical distribution of our procedure is a good approximation to the
% real
% distribution.

\citeasnoun{Luby93} also provide a strategy (a \emph{universal strategy} applicable
to every distribution) to minimize the expected cost of random procedures
in the case where no \emph{a priori} knowledge is available. It
consists of sequences of executions whose values are powers of two.
After two executions with a given threshold, the threshold is changed to its double
value. Let $t_{i}$ be the number of generations of the i-th execution; 
the universal strategy is defined as:
\begin{equation*}
t_{i}=
\begin{cases}
2^{k-1} & \text{if} \ i = 2^k-1 \\ t_{i-2^{k-1}+1}, & \text{if} \
  2^{k-1} \leq i < 2^{k}-1,
\end{cases}
\end{equation*}
yielding strategies of the form
\begin{equation*}
(1,1,2,1,1,2,4,1,1,2,1,1,2,4,8,1,\ldots).
\end{equation*}

\citeasnoun{Luby93} presents two theorems which together prove the
asymptotic optimality of this procedure for an unknown distribution.
% 
% \begin{thm}
% For all distributions $p$,
% \begin{equation}
% E[\mathcal{S}_{\text{Luby}}] \leq
% 192E[\mathcal{S}_{\theta^{*}}](\log_{2}E[\mathcal{S}_{\theta^{*}}]+5).
% \end{equation}
% \end{thm}
% 
% \begin{thm}
% For any strategy $\mathcal{S}$,
% \begin{equation}
% \sup_{p:l_{p}=l}E[\mathcal{S}] \ge
% \frac{1}{8}E[\mathcal{S}_{\theta^{*}}]\log_{2}(E[\mathcal{S}_{\theta^{*}}]).
% \end{equation}
% \end{thm}
% 
% Where $E[\mathcal{S}_{\text{Luby}}]$ is the expected execution time using the
% universal strategy $\mathcal{S}_{\text{Luby}}$, and
% $E[\mathcal{S}_{\theta^{*}}]$ is the average time using a restart strategy
% with optimal fixed threshold $\theta^{*}$. Therefore, it is possible to find a
%probability
% distribution of the execution time for any strategy which makes its expected
%time to be at least
% $O(E[\mathcal{S}_{\theta^{*}}]\log_{2}E[\mathcal{S}_{\theta^{*}}])$,
% while this is the worst case for the strategy
% $\mathcal{S}_{\text{Luby}}$. In expected terms,
% $\mathcal{S}_{\text{Luby}}$ is the best which can be obtained without a priori
%knowledge
% about the distribution of the execution time.

Table \ref{tabla_optima} summarizes the results of the application of both
strategies. The average time using restarts with the
universal strategy is approximately twice the time needed using fixed restarts with the optimal
threshold. Both yield a considerable acceleration against the algorithm without restarts.

\begin{table}
\centering
\begin{tabular}{|r|r|r @{.} l|r|}
\hline dimension & no restart &
\multicolumn{2}{c|}{optimal fixed threshold} &
universal\\ \hline 1.3 & 1,173 & 164 & 9655
$(\theta^{*}=6)$ & 294.867 \\ \hline 1.5 & 1,108 & 374 & 2069
$(\theta^{*}=10)$ & 622.181 \\ \hline 1.8 & 1,443 & 248 & 5263
$(\theta^{*}=17)$ & 625.334 \\ \hline 2 & 5.4360 & 1 & 1655
$(\theta^{*}=1)$ & 1.1701 \\ \hline
\end{tabular}
\caption{A comparison between average execution times for each dimension without
restarts, with an optimal fixed threshold strategy and with the universal
strategy.}
\label{tabla_optima}
\end{table}

In several problems whose execution times had heavy tail distributions,
the universal strategy was found to grow `too slowly.' This happens because, in
those problems, the restart sequence takes too many iterations to reach a
value near $\theta^{*}$ \cite{Cebrian07,Kautz02}.
A correction was proposed by \citeasnoun{walsh99search}, with a new restart
strategy
which was applied successfully to
constraint satisfaction problems. In this simple strategy, each new restart is a
constant factor $\gamma$ greater than the preceding value:
\begin{equation}
(1,\gamma,\gamma^2,\gamma^3,\ldots), \quad
\gamma>1.
\end{equation}

This strategy has a high probability of success when the restart value
$t_{i}=\gamma^{i-1}$
is near the optimal restart threshold value. Increasing  the
restart threshold geometrically
makes sure that the optimal value will be reached in a few generations.
The solution is expected to be found within a few restarts after the value of
$t_{i}$ has surpassed the optimal. This strategy has the advantage of being
less sensitive to the actual distribution it is applied to.

Figure \ref{tabla_walsh} displays the average execution times using
Walsh strategy for several values of $\gamma$. It can be seen that
$\gamma=1.2$ provides the fastest acceleration; with this parameter,
fractal dimension 2 reaches the performance of fixed restarts with optimal
threshold. The average times for fractal dimensions 1.3 and 1.5 are
approximately double of those obtained with the universal strategy, although
much less than those without any restart strategy. A special case is
fractal dimension 1.8, where Walsh
strategy worsens the performance.
% LocalWords:  portfolios Luby Bernoulli Ertel superiormente aleatorizada SAT

\begin{table}[t]
\centering
\begin{tabular}{|r|r|r|r|r|r|r|}
\hline dimension &
\multicolumn{5}{c|}{Walsh} \\ \cline{2-6} &
$\gamma = 1.2$ & $\gamma = 1.4$ & $\gamma = 1.6$ & $\gamma = 1.8$ &
$\gamma = 2$ \\ \hline 1.3 & 441.5138 & 639.0714 & 846.0743 & 773.4603
& 898.4630 \\ \hline 1.5 & 654.5434 & 773.9020 & 845.0908 & 905.0780 &
938.3790 \\ \hline 1.8 & 3,115 & 2,695 & 2,527 & 2,437 & 2,372 \\
\hline 2 & 1.167 & 1.1827 & 1.1729 & 1.1979 & 1.1783 \\ \hline
\end{tabular}
\caption{Average execution times using the Walsh strategy for several values of
$\gamma$.}
\label{tabla_walsh}
\end{table}

%% file: cap4-4-7.tex
\section{Conclusions and future work}

Heavy tail probability distributions have been used to model several real 
world phenomena, such as weather patterns or delays in large communication 
networks. In this paper we have shown that these distributions may be also
suitable to model the execution time of an algorithm which uses Grammatical Evolution for
automatic fractal generation. Heavy tail distributions help to explain the
erratic behavior of the mean and variance of this execution time
and the large tails exhibited by the distribution.

We have proved that restart strategies mitigate the
inconveniences associated with heavy tail distributions and yield a considerable
acceleration on the previous algorithm.
These strategies exploit the non-negligible probability
of finding a solution in short executions, thus reducing the
variance of the execution time and the possibility that the algorithm fails,
which improves the overall performance.

We have given evidence that several restart strategies are of  practical
value, even in scenarios with no a priori  knowledge about the probability distribution
of the execution time.

So far, we have considered situations of complete or inexistent knowledge. In
real situations, the execution time or the resources are bounded, so that
some \emph{partial knowledge} about the execution time is available.
In this scenario, we suspect that our algorithm would take advantage of
\emph{dynamic restart strategies} based on predictive models, which
have been used successfully to tackle decision and combinatorial problems
\cite{horvitz2001bat,Kautz02,ruan2002rpd}. Further research along this line
would be focused on pinpointing the real time knowledge about the behavior of
the algorithm which would make it possible to build predictive models
for its execution time, thus providing a further acceleration.

Finding the conditions for the execution time of a particular Grammatical Evolution
experiment to exhibit a heavy tail distribution would also make an interesting research line:
is the fractal generation optimization exhibiting a typical behavior or just an
exception?

% \citeasnoun{Goldberg1992mmd[29]} provide an example:
% the mean 
% of the execution time is much greater when genetic algorithms are used
% to solve a deceptive problem,  compared to solving a non deceptive problem.

% To overcome these difficulties, some researchers Horvitz et al. [46] have
% designed predictive models
% for the execution time of randomized solvers for the SAT and CSP problems,
% which
% use a Bayesian frame to estimate the time needed to solve the problem as a
% function of a few domain-dependent and domain-independent aspects measured
% during the first steps of the genetic algorithm. This work was extended by 
% Kautz et al. [51], using real time knowledge about the behavior of the solvers
% to 
% design restarts strategies. They have devised an optimal strategy for dynamic 
% restarts, showing its efficiency by means of the empirical results of
% randomized real 
% solvers for two problems: SAT and CSP. A further extension by Ruan et al.
% [96] 
% has considered execution dependencies between different runs with the same 
% subjacent distribution, showing that, in this case, an off-line dynamic
% programming 
% approach could generate the optimal strategy and could be combined with real 
% time information to improve its performance.